\documentclass[conference]{IEEEtran}
\IEEEoverridecommandlockouts
\usepackage{cite}
\usepackage{amsmath,amssymb,amsfonts}
\usepackage{subfigure}
\usepackage{algorithm,algcompatible}
\usepackage{booktabs}
\usepackage{graphicx}
\usepackage{textcomp}
\usepackage{xcolor}
\def\BibTeX{{\rm B\kern-.05em{\sc i\kern-.025em b}\kern-.08em
    T\kern-.1667em\lower.7ex\hbox{E}\kern-.125emX}}
\begin{document}

\title{HQNAS: Auto CNN deployment framework for joint quantization and architecture search\\
{\footnotesize}
\thanks{}
}

\author{\IEEEauthorblockN{ Hongjiang Chen$^1$
, Yang Wang$^1$
, Leibo Liu$^1$
, Shaojun Wei$^1$
, Shouyi Yin$^1$}
\IEEEauthorblockA{\textit{
$^{1}$Institute of Microelectronics, Tsinghua University, Beijing 100084, China.} \\
}
}
\maketitle
\begin{abstract}
Deep learning applications are being transferred from the cloud to edge with the rapid development of embedded computing systems. In order to achieve higher energy efficiency with the limited resource budget, neural networks(NNs) must be carefully designed in two steps, the architecture design and the quantization policy choice. Neural Architecture Search(NAS) and Quantization have been proposed separately when deploying NNs onto embedded devices. However, taking the two steps individually is time-consuming and leads to a sub-optimal final deployment. To this end, we propose a novel neural network design framework called Hardware-aware Quantized Neural Architecture Search(HQNAS) framework which combines the NAS and Quantization together in a very efficient  manner using weight-sharing and bit-sharing. It takes only $\sim$4 GPU hours to discover an outstanding NN policy on CIFAR10. It also takes only $\%$10 GPU time to generate a comparable model on Imagenet compared to the traditional NAS method  with 1.8x decrease of latency and a negligible accuracy loss of only 0.7$\%$. Besides, our method can be adapted in a lifelong situation where the neural network needs to evolve occasionally due to changes of local data, environment and user preference.

\end{abstract}


\section{Introduction}

Recently, the great success of neural networks has been witnessed in various challenging computer vision tasks, e.g, image classification\cite{ref1}, object detection\cite{ref2} and semantic segmentation\cite{ref3}. Transferring the deep learning applications from the cloud to edge is becoming a trend with the rapid development of embedded computing systems. However, deploying a  neural network onto an edge device is still a challenging work, which usually takes two laborious steps. Firstly, the architecture of the network must be carefully designed for a specific hardware platform. Secondly, the network should be further compressed for the sake of energy efficiency. Both of these process are time-consuming and demand much expert knowledge.

\begin{figure}[htbp] 
\centering 
\includegraphics[width=0.52\textwidth]{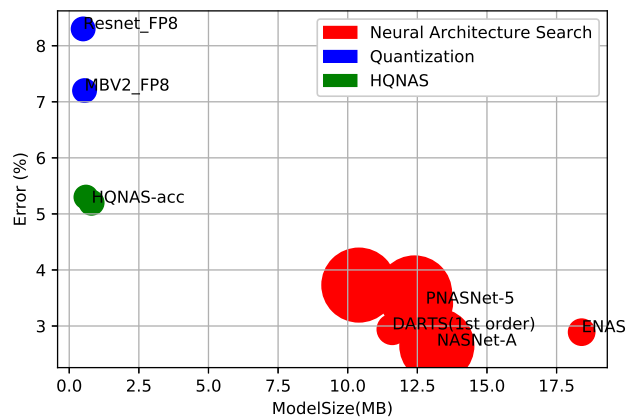} 
\caption{Model camparisons of different NAS and Quantization framework, and the size of the circle represents the corresponding relative computation resource.} 
\end{figure}

For the first step, great research interests have been raised in developing auto machine learning(AutoML) tools, among which Neural Architecture Search(NAS) is the shiniest aiming at finding high-performance networks. Most of these methods are based on Reinforcement Learning (RL) and Evolution Algorithm (EA). A typical RL-based NAS methods usually build networks by a RNN controller\cite{ref4} to find a sequence of operator and connection choice. For the second step, pruning, sparse training and quantization\cite{ref5,ref6,ref7} are three common ways to further compress the model.

However,  taking the two steps individually is intricate and leads to a sub-optimal final deployment. Besides,  once the whole process is finished, the network cannot be changed even when new demands appear. Based on this, we propose a novel neural network design framework called Hardware-aware Quantized Neural Architecture Search(HQNAS) framework which combine the NAS and Quantization together in a very efficient  manner using weight-sharing and bit-sharing. We also utilize the Pareto boundary\cite{ref10} to adjust the multi-objective function so that the model can meet the demands of different accuracy and latency. During the searching process,  the architecture and the quantization policy will be determined as long as the training for the super-net is  finished. After that, the generated simplified network will be trained again while the all the data in the super-net will be preserved for the next network evolution.

Our contributions of this work can be concluded as follows:
\begin{itemize}
\item We propose a fast end to end neural network deployment framework which can implement
 NAS and Quantization at the same time.
\item We utilize a semi-regressed Pareto function to yield hardware-aware models and make the whole process efficient by means of weight-sharing and bit-sharing.
\item We demonstrate that our method can be adapted in a lifelong situation where the neural network needs to evolve occasionally due to changes of local data, environment and user preference.
\end{itemize}

\section{Related Work}

\subsection{Neural Architecture Search}
Great research interests have been raised in auto machine learning. NAS is one of these tools aims at automatically designing network. Current research usually fall into three categories: RL, EA and DARTS\cite{ref8}. Typically, RL-based models compress the information of the architecture of NNs into a well-regulated sequence by training a RNN controller. Where EA-based models generate architectures by means of evolutionary and mutation algorithm. Both RL-based and EA-based methods achieve comparable results with human-designed networks.

However,  RL-based and EA-based methods are quite time-consuming which usually take more than 100 gpu hours to run a complete search process. Hence a new method called DARTS have been applied which regresses the original discrete search space into continuous valuables. Methods based on DARTS searches over an over-parameterized “superkernel” in each layer to optimize for the best of convolution kernel weights. Although these methods are more efficient than those based on RL and EA, it still suffers from the memory explosion problem introduced by the multi-path supernet\cite{ref9}.

\subsection{Model compression}
Several model compression methods are also proposed for higher energy efficiency, e.g, network pruning\cite{ref5}, distillation\cite{ref6} and special convolution structures\cite{ref7}. Quantization is also a significant branch used in real applications \cite{ref8}. It can effectively save storage space and communication cost by reducing the model size. Regardless of the different layers’ redundancy, current methods take the uniform precision for the entire network. It seems more promising to determine mixed precision quantization policy which is supported by a lot of hardware platforms, e.g., CPUs and FPGAs. It is diffuse to finish the job by human experts, since a typical model has more than tens or hundreds of layers.

\section{approach}
In this section, we introduce our entire framework, which combines the NAS and the quantization based on the single-path DARTS\cite{ref9}. By using weight-sharing and bit-sharing, the whole process is efficient and can generate comparable models. 
\subsection{Problem Definition}
A desired model s=$\{\alpha,\theta\}$ is consists of the architecture 
$\alpha$ and the quantization policy $\theta$. We difine our problem as a muti-objective optimization  and take use of a Pareto reward function $\mathcal{R}(s)$ to search for the best model:
\begin{equation}
s=\mathop{\arg\max}_{s}\mathcal{R}(s)
\end{equation}
\begin{equation}
\mathcal{R}(s)=-\mu Lat(s)+\nu Acc(s)
\end{equation}

Where  \textit{Lat(s)} is the latecncy of the model running on a specifiac platform and \textit{Acc(s)} is the accuracy on the validation data set. And ($\mu,\nu$) are the weighting factors for the latency and the accuracy where $\mu+\nu$=1. By adjusting values of $\mu$ and $\nu$, we can change the trade-off between the accuracy and the runtime.

The whole framework  focuses on finding a neural architecture $\alpha$ and a quantization policy $\theta$ to build the optimal model s=$\{\alpha,\theta\}$. In our experiment, we soften the optimization problem when desired model has threshold for latency or accuracy as follows:

\begin{equation}
Lat(s)=\left\{
\begin{aligned}
&Lat(s)-Lat_{the}  ,\  if \ Lat(s)<Lat_{the}\\
&0.5*(Lat(s)-Lat_{the}), \ otherwise
\end{aligned}
\right.
\end{equation}

Where $Lat_{the}$ is the threshold of the optimization condition for latency. The softened function is inspired by the similar form of Leaky-Relu\cite{ref12} in order to guide the searching process on qualified models.

\subsection{Search Space}
\begin{figure}[htbp] 
\centering 
\includegraphics[width=0.5\textwidth]{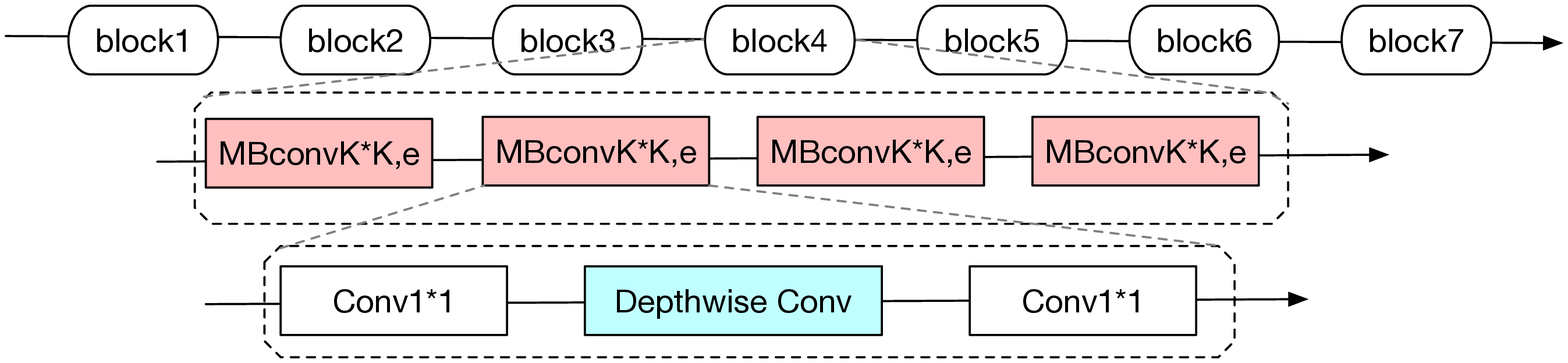} 
\caption{The whole architecture is based on MobileNetV2 consits of 7 blocks(5 for CIFAR10), each depthwise convolution layer is searchable.} 
\end{figure}
We separate the whole search space into neural architecture search space and quantization search space, $S=\{S_{\alpha},S_{\theta}\}$. 

For neural architecture search space, our framework builds upon hierarchical MobileNetV2-like\cite{ref12} search spaces. The main goal is to deside the type of mobile inverted bottleneck convolution. To be more specific, MBConv layer is decided by  the kernel size of the depthwise convolution k, and by expansion ratio e. In particular, we consider MBConv layers with kernel sizes {(3, 5)} and expansion ratios of {(3, 6)}.

For quantization policy search space, we focus on the optimal policy for the weight values. As showed in fig.2, we bulid the whole framwork based on the MobileNetV2 consists of several blocks. Where each block consists of 4 groups of a point-wise 1*1 convolution, a k*k depthwise convolution, and a linear 1*1 convolution(1 group for the first and the last). The backbone has 5 blocks for CIFAR10 dataset and 7 blocks for Imagenet. To reduce the complexity of the problem, we asume each layer within a group share the same quantization policy. The choices of quantization are limited within (4, 8, 16) bits.

The total number of architectures and quantization can be caculated as $N_a=4^{3*4+1+1}$, $N_q=3^{3*4+1+1}$ for a 5 blocks backbone. Thus the total search space is $N_a*N_q\simeq1.28*10^{15}$.

\subsection{Weight-sharing and Bit-sharing}
\begin{figure}[htbp] 
\centering 
\includegraphics[width=0.38\textwidth]{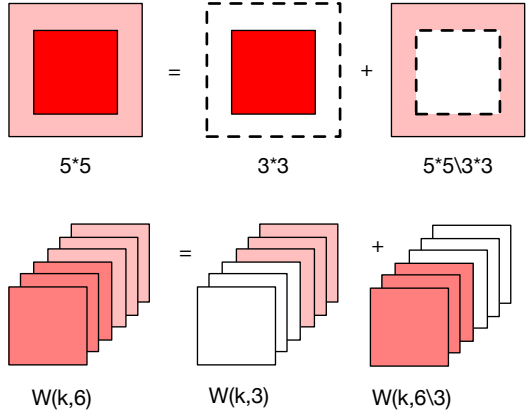} 
\caption{The idea of weight-sharing is to choose how much part of the super-kernel to generate a new one. Each new kernel shares the same super-kernel.} 
\end{figure}
Weight-sharing: The key idea of weight-sharing is relaxing NAS decisions over an over-parameterized kernel. A super-kernel is the max size of the potential kernel. For MobilenetV2, it is a $5*5$ kernel with the expansion ratio of 6. As showed in fig.3, we divide the whole kernel in several parts.
\begin{itemize}
\item For searching kernel size, choosing $3*3$ or $5*5$ is equivalent to whether choose the subset of $w_{55\backslash33}$.

\begin{equation}
w_k=w_{33}+\sigma(id(w_{55\backslash33},t_{k,5}))*w_{55\backslash33}
\end{equation}

\item For searching expand ratio and skip opertaion, it is determined by if we choose to preserve the entire length, half length or zero of the kernel.

\begin{equation}
w=\sigma(id(w_{k,3},t_{e,3}))*(w_{k,6\backslash3}+\sigma(id(w_{k,3},t_{e,6}))*w_{k,6\backslash3})
\end{equation}

\end{itemize}

Where three $\sigma(\cdot)$ are the sigmoid functions used to decide whether we choose the subset of $w_{55\backslash33}$, the skip operation, and the expand ratio of 3.

The original problem is a hard-decision problem which can be softened by using sigmoid functions. The $id(\cdot)$ function is used to caculate the indicator value as input of the sigmoid. We simply use the group Lasso term of the corresponding weight values and a threshold to make our decisions.
\begin{equation*}
id(w_{55\backslash33},t_{k,5})=\parallel w_{55\backslash33}\parallel^2-t_{k,5}
\end{equation*}

\begin{figure}[htbp] 
\centering 
\includegraphics[width=0.38\textwidth]{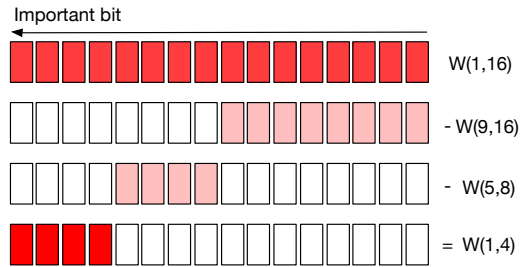} 
\caption{The idea of bit-sharing is to choose whether to leave out the non-important bits.} 
\end{figure}

Bit-sharing: As showed in fig.4, the basic idea of bit-sharing is similar to weight-sharing. The quantization policy is decided by whether we choose to preserve the 5$\sim$8 bit and the 9$\sim$16 bit.
\begin{equation}
\begin{split}
w_q^i=
&w_{16}^i-\sigma (id(t_{9\sim16}^q,w_{9\sim16}^i)) \\
&*(w_{9\sim16}^i+\sigma (id(t_{5\sim8}^q,w_{5\sim8}^i))*w_{5\sim8}^i)
\end{split}
\end{equation}

The inspiration is drawn from the weight-based conditions for quantization-related decisions\cite{ref13}.It is worth noting that the decision of quantization uses the manner  of leaving out which is different from the architecture search. Hence, the order of Lasso and the threshold is inverted.

\begin{equation*}
id(t_{9\sim16}^q,w_{9\sim16}^i)=t_{9\sim16}^q-\parallel w_{9\sim16}^i\parallel^2
\end{equation*}

It is worth mentioning that all the decision made by the sigmoid function is not determinate. We sample different architectures and quantization policy probabilistically during the search. For example, if $\sigma(id(w_{55\backslash33},t_{k,5}))$=0.8, we have 80\% chance to generate a 5*5 kernel when sampling. However when the search is finished, we degenerate the sigmoid to 0-1 function which means we finally get a determinate 5*5 kernel in this case.

\subsection{Quantization Policy}
We fomulate the quantization process as follows:
\begin{equation}
w^*=Norm^{-1}(Q(Norm(w), b))
\end{equation}

Where $(w, w^*)$ is the weight vector of the real value and the approximate quantization value, and b is the bit number. \textit{Norm(w)} is a linear scaling functiong which normalizes the values of the vectors into [0,1] and $Norm^{-1}$ is the inverse function.

In paticular, Q is the quantization function which takes in the normalized real value w and the bit number w. The ouput is the closest quantization value of the corresponding weight element $w_i$:

\begin{equation}
Q(w_i,b)=\frac{\lfloor w_i2^b \rfloor}{2^b}+\frac{\zeta_i}{2^b}
\end{equation}

Where $\zeta_i$ is the rounding factor of each element
\begin{equation*}
\zeta_i=\left\{
\begin{aligned}
&1  ,\  if \ w_i2^b-\lfloor w_i2^b \rfloor>0.5\\
&0, \ otherwise
\end{aligned}
\right.
\end{equation*}

\subsection{Search Strategy}

\begin{algorithm}
\caption{HQNAS} 
\label{alg1}
\begin{algorithmic}
\REQUIRE  training set $D_{train}$, validation set $D_{val}$ \\ max num epochs $E_{max}$, weighter factor $\mu$ and $\nu$ \\ threshold $Lat_{the}$ and $Acc_{the}$, historical checkpoint $M^*$

\ENSURE Model P with architecture $\alpha$ and quantization $\theta$
\IF{$M^*$ exists} 
\STATE $M \Leftarrow innitialize(M*)$  
\ELSE 
\STATE $M \Leftarrow innitialize(w_{random}, theshold_{random})$ 
\ENDIF 
\WHILE {$E<E_{max}$}
\STATE $w \leftarrow w - \nabla_wCE_{train}(M)$
\STATE Sample N Models according to id($\cdot$) function
\STATE $theshold \leftarrow  theshold-\nabla_{theshold}R_{val}(M)$ \\$\quad\quad\quad\quad\approx\frac{1}{N}\sum_{i=1}^{N}R_{val}(Model_i) \nabla_{theshold}log(p(\alpha_i,\theta_i))$ 
\STATE $M \leftarrow (w,theshold)$
\STATE $E \leftarrow E+1$
\ENDWHILE
\STATE $\alpha, \theta \leftarrow determinate\_id(w,theshold)$
\STATE $P \leftarrow train\_M\_with(\alpha,\theta)$
\STATE $M^* \leftarrow (w,theshold)$

\end{algorithmic}
\end{algorithm}

The entire framwork combines the single-path DARTS and REINFORCE to update the weights and the thresholds of the whole network. Commonly, we innitialize the model with random weights and thesholds. When it comes to the life-long learning situation, the network needs to evolve again because of the new local data or the change of the environment and user preference. It is time-consuming to start from scratch to do all the search again. We take use of the historical data checkpoint to accelerate the evolving process.

During the main loop, we update the weight values and the thesholds alternately until reaching the maxepochs. The weights is updated by the cross-entropy caculated on the training set. When it comes to updating the thresholds. Since the latency is not derivable with respect to the thresholds, we adapt the REINFORCE algorithm by sampling N models(architecture and quantization) from the sigmoid function on current weights and thesholds.

When reaching the max epoch, the architecture $\alpha$ and the quantization $\theta$ are determined by the sigmoid function on the final weights and thesholds. Instead of sampling the model with probability, the sigmoid here is degenerated to a 0-1 function. Hence the framework can generate the final model by training the network with the determined $\alpha$ and $\theta$. Finally, we save the previous model as a checkpoint for accelerating the next evolution in future.

\begin{table*}[htbp]
\caption{The results of NAS and quantization policy search methods on CIFAR10}
\normalsize
\centering
\begin{tabular}{ccccccc}
\toprule  
Model& GPUs & Days &Size(MB) &Error(\%)& Method & Quantization Support\\
\midrule  
PNASNet-5\cite{ref14}& 100 &1.5 &12.8 &$3.41\pm0.09$ &SMBO& no\\
NASNet-A\cite{ref15}&500&4&13.2&2.65& RL& no\\
NASNet-B\cite{ref15}&500&4&10.4&3.73& RL& no\\
NASNet-C\cite{ref15}&500&4&12.4&3.59& RL& no\\
ENAS\cite{ref16}&1&0.5&18.4&2.89&RL &no\\
DARTS(1st order)&1&1.5&11.6&2.94& Gradient& no\\
DARTS(2nd order)&1&4&13.6&$2.83\pm0.06$& Gradient& no\\
JASQNet\cite{ref17}&1&3&2.5&2.9&EA+RL&yes\\
\midrule
HQNAS(accuracy prior)&1&0.3&0.8(quantized)&+0.7(compared to MBV2)&Gradient&yes\\
HQNAS(latency prior)&1&0.3&0.6(quantized)&+0.9(compared to MBV2)&Gradient&yes\\

\bottomrule 
\end{tabular}

\end{table*}

\section{experiment}
We use single-path DARTS NAS and REINFORCE algorithm to search for the optimal neural architecture and quantization policy on the data set CIFAR10 and Imagenet. We use CPU and GPU for the target hardware platform. The latency is measured on Tesla V100 with batch size of 8, while the CPU latency is measured on Intel(R) Xeon(R) Platinum 8168 with a batch size of 1. Where CPU latency is used on the CIFAR10 experiment for proving the effectiveness of the joint optimization, while GPU latency is taken as a comparison object for hardware-aware search with results on CPU.

Our backbone architecture is based on MobileNetV2. The whole network consists of 5 blocks for CIFAR10 (7 for Imagenet) and each block is made up of 4 mobile inverted bottleneck convolution layers(1 for the first and the last block). The choices of architecture operations and quantization levels are shown below:
\begin{itemize}
\item Architecture: (3*3, 5*5) kernel size, expand ratio of (3,6)
\item Quantization: (4, 8, 16) bit
\end{itemize}

\subsection{Experiments on CIFAR10}
\begin{figure}[htbp] 
\centering 
\includegraphics[width=0.5\textwidth]{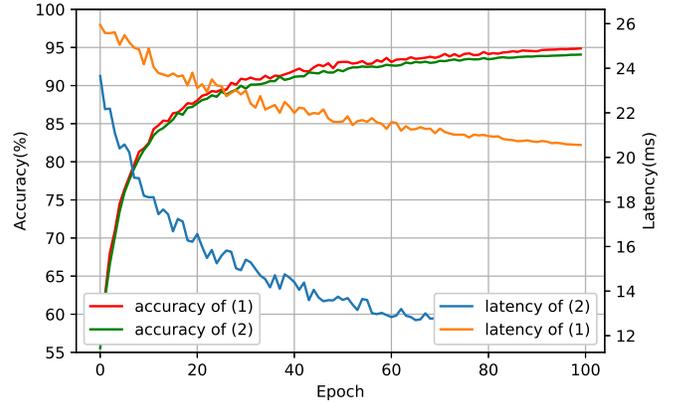} 
\caption{Running curves of the accuracy and the lantency on CIFAR10 experiment (1)for accuracy prior (2)for latency prior} 
\end{figure}

The backbone network architecture is based on a 5-block MobileNetV2. We randomly samples half of the training set as our validation set. The max epoch $E_{max}$ is set to 100, with batchsize 256 for both training and validation sets. We use batch-specific statistics for batch normalization instead of the global moving average to avoid the problem causing by architecture changing during the search process.The number is much bigger than that in DARTS because single-path can reduce the memory explosion problem. Momentum SGD is adapted to optimize the weights in our experiments, with initial learning rate $\alpha=0.05$, momentum 0.9, and weight decay $3*10^{-4}$. We conduct two groups of experiments with different weighted factors for the reward of latency and accuracy:
\begin{itemize}
\item Weighted factors: \\
(1) $\mu=0.2, \nu=0.8$,\quad accuracy-prior\\
(2) $\mu=0.8, \nu=0.2$,\quad latency-prior
\item Qualified theshold:\\ $Lat_{the}=20ms$ for CPU\\
$Acc_{the}=96.8\%$
\end{itemize}

As showed in fig.5, we record the curve of both the validation accuracy and the latency through the searching process. Both of two experiments is finished within 4 hours using a single GPU. During the first 20 epochs, dropout is employed across the different subsets of the kernel weights which is a common technique in order to prevent the supernet from learning as an ensemble\cite{ref9}. After this period, all the curves 
tend to converge in different speed. The curve of accuracy in experiment(1) tends to converge much faster than the other, while in experiment(2) the curve of latency behaves in the same way. This is because the weighted factors guide the framework searching towards what matters most.

\begin{figure}[htbp] 
\centering 
\includegraphics[width=0.5\textwidth]{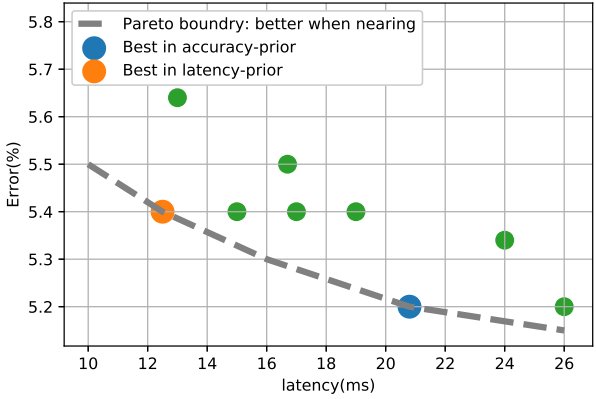} 
\caption{Best models on CIFAR10 are scattered near the Pareto boundry} 
\end{figure}

We record the latest 10 generated models before the searching finishes and scatter them in one figure. It is interesting that the most models are closing to the Pareto boundry while differ in the specific position.

The detail of our framework is listed in table.1 compared with other classic methods. HQNAS is able to match or even outperform the state-of-the-art architecture search methods on image classification problems, while being able to finish the quantization at the same time.


\subsection{Experiments on Imagenet}
\begin{table}[H]

\begin{tabular}{cccc}
\toprule  
Model& Top-1(\%)&  GPU lat(ms) &CPU lat(ms)\\
\midrule  
Proxyless NAS-GPU\cite{ref18}& 75.1& 5.1 &204.0\\
Proxyless NAS-CPU\cite{ref18}& 75.3& 7.4&134.8\\
HQNAS-GPU(Quantized)&74.4& 2.6&128.4\\
HQNAS-CPU(Quantized)&74.1& 3.2&67.6\\
\bottomrule 
\end{tabular}
\caption{Hardware prefers specialized networks}
\end{table}
We change the number blocks from 5 to 7 when it comes to the Imagenet. So total search space can be caculated by $N=4^{1+4*5+1}*3^{1+4*5+1}\simeq5.5*10^{23}$. We conducte the whole search twice with GPU and CPU as latecny feedback platform respectively. The validation set is constructed by sampling 50000 images from the training set. The hyperparameters of the framework are carefully choosen as: 
\begin{itemize}
\item Weighted factors: $\mu=\nu=0.5$
\item Qualified theshold:\\ $Lat_{the}=5.0$ ms for GPU, 140 ms for CPU\\
$Acc_{the}=72.0\%$
\end{itemize}

\begin{figure}[htbp] 
\centering 
\includegraphics[width=0.5\textwidth]{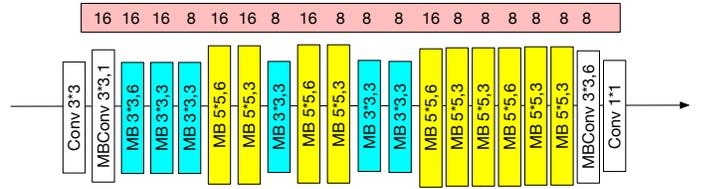} 
\caption{Best model generated for the GPU platform} 
\label{Fig.main2} 
\end{figure}

\begin{figure}[htbp] 
\centering 
\includegraphics[width=0.5\textwidth]{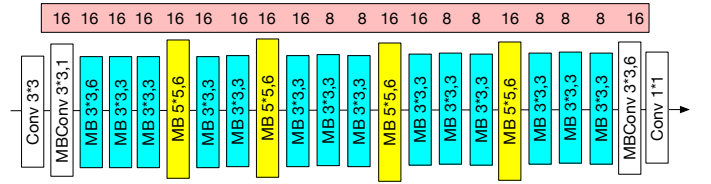} 
\caption{Best model generated for the CPU platform} 
\label{Fig.main2} 
\end{figure}

As showed in table.2, our results are comparable with similar work both on GPU and CPU. Compared to Proxyless NAS with float weight number. Our joint optimization greatly reduce the latency with a negligible accuracy loss of only 0.7\%$\sim$1.2\% but $\sim1.8$× fewer latency. The whole search time is 100× shorter than previous NAS works.

The final network architecture and quantization are showed in fig.7 and fig.8. It is worth noting that the architecture and the quantization policy are quite different on two hardware platforms. We have similar findings that the GPU model is shallower and wider than CPU model because GPU has higher parallelism. While GPU model prefers shorter bit lenth especially in last several layers. And both have quite shallow layers in early stage because the early feartures is relatively small and need higher resolution. Besides, both GPU model and CPU model prefer larger MBConv and wider quantization bit in the downsampled layer.

\subsection{Lifelong learning Situation}

In this part, we demonstrate that our framework can be adapted to life long learning situations. The typical NN deployment framework ususlly freezes the network once it is transferred to embedded device, which is not reasonable as the local data set may increase(e.g. new labels). As a result, the NN needs to learn from the incremental knowledge and evolve itself. Deploying the entire NN starting from scratch with the new data is feasible but inefficient.
\begin{figure}[htbp] 
\centering 
\includegraphics[width=0.48\textwidth]{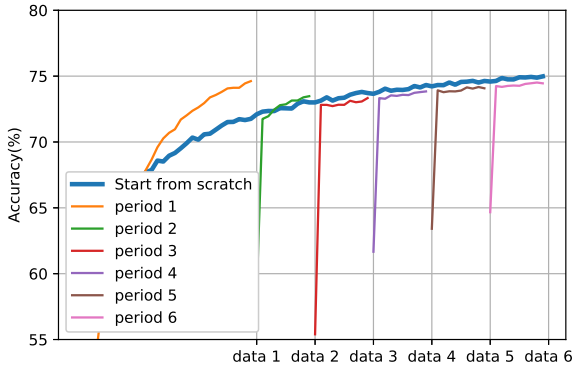} 
\caption{Different running curves of training from scratch and using HQNAS with historical checkpoint} 
\label{Fig.main2} 
\end{figure}
We construct our experiment by randomly select 500 catagories of Imagenet as the original dataset and use HQNAS to search for the architecture and the quantization. Then we add 100 labels of data each time into a new set of data and directly search with the hitorical check point of HQNAS untill 1000 catagories. We record the difference between the incremental learning curve and the curve learning from scratch to fit for 1000 catagories.

As showed in fig.9, although the total time costed by the incremental HQNAS is a little more than starting from scratch. The average time used by each update is quite limited, which matters in our real life. By observing the weight values of a fixed layer, we find the weight vector almost unchanged if the local data does not get too much new knowledge at a time, which is natural in real life. Hence, our HQNAS framework is promising in most of those life long learning situations with slowly incremental knowledge.
\begin{figure}[htbp] 
\centering 
\includegraphics[width=0.5\textwidth]{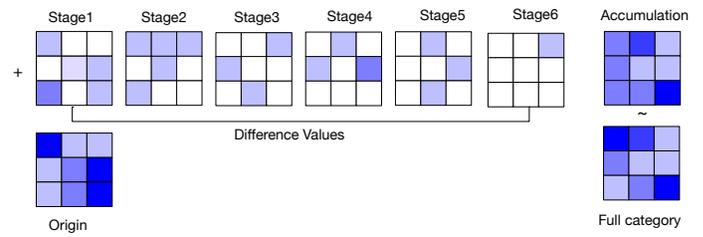} 
\caption{Each time the weight values do not change a lot if not taking in too much knowledge at 1 time} 
\label{Fig.main2} 
\end{figure}

\section{conclusion}
In this paper, we presents HQNAS which combines neural architecture search and quantization together. By using weight-sharing and bit-sharing, HQNAS can search for a qualified model very efficiently. Based on single-path DARTS and REINFORCE algorithm, the generated model is quite comparable with those designed by human experts while being convenient to acquire. Besides, HQNAS presents the potential in life-long learning situations such as incremental learning where the local data increase without losing hitorical work. The ability to evolve fast and automatically is very promising to be applied on embeded devices in future. For our following research, we would like to investicate the combination of NAS with other compression methods(e.g. sparsity) and enlarge the architecture search to other wondrous networks.


\begin{thebibliography}{99} 
\bibitem{ref1} K. He, X. Zhang, S. Ren, and J. Sun. Deep residual learning for image recognition. In CVPR, pages 770–778, 2016.
\bibitem{ref2} S. Ren, K. He, R. B. Girshick, and J. Sun. Faster R-CNN: towards real-time object detection with region proposal net- works. IEEE Trans. Pattern Anal. Mach. Intell., 39(6):1137– 1149, 2017. 
\bibitem{ref3} L. Chen, G. Papandreou, I. Kokkinos, K. Murphy, and A. L. Yuille. Deeplab: Semantic image segmentation with deep convolutional nets, atrous convolution, and fully connected crfs. IEEE Trans. Pattern Anal. Mach. Intell., 40(4):834– 848, 2018.
\bibitem{ref4}B. Zoph and Q. V. Le. Neural architecture search with rein- forcement learning. CoRR, abs/1611.01578, 2016.
[35]
\bibitem{ref5}S. Han, H. Mao, and W. J. Dally. Deep compression: Com- pressing deep neural network with pruning, trained quanti- zation and huffman coding. CoRR, abs/1510.00149, 2015.
\bibitem{ref6}J. Ba and R. Caruana. Do deep nets really need to be deep? In NIPS, pages 2654–2662, 2014.
\bibitem{ref7}A. Polino, R. Pascanu, and D. Alistarh. Model compression via distillation and quantization. CoRR, abs/1802.05668, 2018.
\bibitem{ref8}H. Liu, K. Simonyan, and Y. Yang. DARTS: differentiable architecture search. CoRR, abs/1806.09055, 2018.
\bibitem{ref9}Stamoulis D, Ding R, Wang D, et al. Single-path nas: Designing hardware-efficient convnets in less than 4 hours[C]//Joint European Conference on Machine Learning and Knowledge Discovery in Databases. Springer, Cham, 2019: 481-497.
\bibitem{ref10}K. Deb. Multi-objective optimization. In Search methodolo- gies, pages 403–449. 2014.
\bibitem{ref11}Sandler M, Howard A, Zhu M, et al. Mobilenetv2: Inverted residuals and linear bottlenecks[C]//Proceedings of the IEEE conference on computer vision and pattern recognition. 2018: 4510-4520.
\bibitem{ref12}Zhang X, Zou Y, Shi W. Dilated convolution neural network with LeakyReLU for environmental sound classification[C]//2017 22nd International Conference on Digital Signal Processing (DSP). IEEE, 2017: 1-5.
\bibitem{ref13}Ding, R., Liu, Z., Chin, T.W., Marculescu, D., Blanton, R.: Flightnns: Lightweight quantized deep neural networks for fast and accurate inference. In: 2019 Design Automation Conference (DAC) (2019)
\bibitem{ref14}C. Liu, B. Zoph, M. Neumann, J. Shlens, W. Hua, L. Li, L. Fei-Fei, A. L. Yuille, J. Huang, and K. Murphy. Progres- sive neural architecture search. In ECCV, pages 19–35, 2018.
\bibitem{ref15}B. Zoph, V. Vasudevan, J. Shlens, and Q. V. Le. Learn- ing transferable architectures for scalable image recognition. CoRR, abs/1707.07012, 2017.
\bibitem{ref16}H. Pham, M. Y. Guan, B. Zoph, Q. V. Le, and J. Dean. Ef- ficient neural architecture search via parameter sharing. In ICML, pages 4092–4101, 2018.
\bibitem{ref17}Chen Y, Meng G, Zhang Q, et al. Joint neural architecture search and quantization[J]. arXiv preprint arXiv:1811.09426, 2018.
\bibitem{ref18}Cai H, Zhu L, Han S. Proxylessnas: Direct neural architecture search on target task and hardware[J]. arXiv preprint arXiv:1812.00332, 2018.
\end{thebibliography}
\end{document}